\documentclass[times,twocolumn,final,authoryear]{elsarticle}

\usepackage{prletters}
\usepackage{framed,multirow}
\setcitestyle{square}

\usepackage{amssymb}
\usepackage{latexsym}

\usepackage{url}
\usepackage{xcolor}
\usepackage{amsmath}
\DeclareMathOperator*{\argmax}{arg\,max}

\definecolor{newcolor}{rgb}{.8,.349,.1}

\begin{document}

\thispagestyle{empty}

\clearpage

\ifpreprint
  \setcounter{page}{1}
\else
  \setcounter{page}{1}
\fi

\begin{frontmatter}

\title{3DPyranet Features Fusion for Spatio-temporal Feature Learning}

\author[1,2,3]{Ihsan Ullah \corref{cor1}} 
\cortext[cor1]{Corresponding author: }
\ead{ihsan.ullah@universityofgalway.ie}

\author[1]{Alfredo Petrosino}

\address[1]{CVPR Lab, University of Naples 'Parthenope', Naples, 80143, Italy}
\address[2]{Department of Computer Science, University of Milan, Milan, Italy}
\address[3]{School of Computer Science, University of Galway, Galway, Ireland}


\begin{abstract}
Convolutional neural network (\textit{CNN}) slides a kernel over the whole image to produce an output map. This kernel scheme reduces the number of parameters with respect to a fully connected neural network (\textit{NN}). 
While \textit{CNN} has proven to be an effective model in recognition of handwritten characters and traffic signal sign boards, etc. 
recently, its deep variants have proven to be effective in similar as well as more challenging applications like object, scene and action recognition.
Deep \textit{CNN} add more layers and kernels to the classical \textit{CNN}, increasing the number of parameters, and partly reducing the main advantage of \textit{CNN} which is less parameters.
In this paper, a 3D pyramidal neural network called \textit{3DPyraNet} and a discriminative approach for spatio-temporal
feature learning based on it, called \textit{3DPyraNet-F}, are proposed.
\textit{3DPyraNet} introduces a new weighting scheme which learns features from both spatial and temporal dimensions analyzing multiple adjacent frames
and keeping a biological plausible structure. It keeps the spatial topology of the input image and presents fewer parameters and 
lower computational and memory costs compared to both fully connected \textit{NNs} and recent deep \textit{CNNs}.
\textit{3DPyraNet-F} extract the features maps of the highest layer of the learned network, fuse them in a single vector, and provide it as input in such a way to a linear-SVM classifier that enhances the recognition of human actions and dynamic scenes from the videos. 
Encouraging results are reported with \textit{3DPyraNet} in real-world environments, especially in the presence of camera induced motion. Further, \textit{3DPyraNet-F} clearly outperforms the state-of-the-art on three benchmark datasets and shows comparable result for the fourth.

\end{abstract}



\end{frontmatter}



\section{Introduction}
\label{sec1:intro}
Recognition tasks such as human actions (e.g. Running, Walking, Clapping, etc.) as well as dynamic natural scenes for example Beach and/or Fire in the videos is an important and highly researched area of computer vision (\textit{CV}) and machine learning (\textit{ML}).
Previously, \textit{CV} approaches use to extract spatio-temporal features with traditional handcrafted descriptors. These features are then classified with a state-of-the-art classifier or SVM. On the other hand, \textit{ML} algorithms learn discriminative features automatically from the given training data. The trained model is used to give accurate predictions for the presented testing data.
On one side, handcrafted descriptors used in \textit{CV} such as \textit{HoF} or \textit{HoG} computed on \textit{STIP} \cite{Laptev2007}, shows good performance for human action recognition (\textit{PyraNet}), but not for dynamic scenes. 
On the other side, combinations of \textit{HOF+GIST} or \textit{MSOE} \cite{Theriault2013,Derpanis2012} or many others show good result for dynamic scene recognition/understanding (\textit{DSR / DSU}) but not for \textit{AR}. 
No model exists equally effective for both tasks.
\newline
In general for human, recognizing an action or dynamic scene in a single image is hard compare to a video. Similarly, even though there is significant advancement being made in image classification algorithms, an action or dynamic scene are hard to be recognized from a single frame due to the reason that a single image/frame don't have enough information. 
Therefore, using temporal information -- that is utilizing continuous frames -- can improve the performance of classifiers, however it still faces challenges because motion is often connected with several artifacts: lighting, specular effects, motion in videos due to the abrupt movement of camera handling and more. Although, there are several models that target individual or some of the problems, however, few well-known temporal models are presented in  \cite{Schindler2008,Yang2014,Liu2015,Melfi2013,Ji2013a,Efros2003,Schuldt2004,Ballan2012,feichtenhofer2013spacetime,feichtenhofer2014bags}. 
These, as well as other papers, claim 90+\% accuracy on specific datasets under their respective targeted scenarios. In real world, this performance is overly optimistic, as humans and scenes change dramatically from one frame to the next, varying in pose, occlusions, illuminations and interactions with the surrounding environment.
\newline
Most current research in human \textit{AR} \cite{Taylor2010,Freitas2010,Le2011,Ji2013a,Krizhevsky2012b} and \textit{DSR} \cite{Tran2014a,Karpathy2014} uses building a deep neural networks (\textit{DNN}) for learning more discriminative and flexible features. These new end-to-end learning models became popular as a result to their good performance on pretty big datasets \cite{trecvid.features,Krizhevsky2012b}. Further, another reason for their wide usage is that the same model with slight tuning can perform well for both \textit{AR} and \textit{DSR} \cite{Tran2014a,Karpathy2014}. 
An important aspect of convolutional \textit{DNN} models is weight learning and sharing concept \cite{Lecun1998}. 
Parameters in a convolutional model are arranged in a kernel, that is not specific to any neuron, rather it is slided and shared over the whole image. Compared to conventional fully connected (FC) \textit{NN} models, this schema reduces the number of parameters, but also increases the chance to reduce their discriminative power while considering the huge amount of data from the videos.
\newline
%
Classical \textit{CV} approaches use a coarse to fine refinement approach, recent deep learning (\textit{DL}) models do not. The idea underlying the refinement is to build a hierarchy of features -- a pyramid -- and to finally select the most discriminative ones for
the classification step.  
This process mimics how the human brain processes images and is considered to be biologically plausible.
An image pyramid decreases the resolution of an image exponentially at each higher layer and it has been successfully used in different
feature extractors/descriptors, e.g. Steerable/Laplacian pyramids \cite{BEIL1994453,1095851}, SIFT \cite{Lowe2004}, SPM
\cite{Lazebnik2006,Jia} and more. 
In the past, most models were pyramidal and were following this biological plausible structure \cite{Cantoni2002,Phung2007,Fukushima1988}. 
Contrarily, these recent \textit{DNN} are not following a coarse to fine refinement of features or the strict biological plausible pyramidal structure. This results in an having large amount of features in final layer, increase in convergence time, ambiguity, and the number of parameters.
\newline
The current focus of DL research for enhancing \textit{CNN} models can be divided in: increasing number of layers (depth) \cite{Krizhevsky2012b,Simonyan14c,SzegedyLJSRAEVR15,He2015c}, increasing number of kernels/maps at each layer (width) \cite{Simonyan14c, ZeilerF14}, introducing/enhancing activation functions (\textit(AF)) \cite{ZeilerStochastic, HeZRS15}, avoiding vanishing gradients despite the depth \cite{SzegedyLJSRAEVR15, Lee2014, Huang2016}, weight initialization \cite{maas2013rectifier, Han2015}, finding impact/structure of kernel size (receptive field (RF)) \cite{Simonyan14c}, introducing/enhancing operations at each step (convolution and pooling) \cite{Graham14a}, reducing large number of parameters, and network structure \cite{NINLinCY13,He2015c,Pang2016,Huang2016}. 
We introduce a new network structure and weighting scheme that focuses on the last five areas. 
Our contributions are: i) A new model that uses spatial and temporal information and a purposely designed weighting schema is proposed called 3D pyramidal Neural Network (\textit{3DPyraNet}), ii) The weighting schema is more suitable for learning the features from videos containing additional abrupt movement due to the camera, iii) Features from trained \textit{3DPyraNet} are fused and further given for training and classification with a linear-\textit{SVM} classifier, and iv) The extension, called \textit{3DPyraNet-F}, can be applied in a wide spectrum of applications (with slight tuning), not needing handcrafted features.
The paper is organized as follows: Section \ref{sec:2dpyranet} provides an explanation and key points of \textit{PyraNet}. Section~\ref{sec:3dpyranet} gives a motivational background of the proposed model. Further, its sub-sections give details about the models that are utilized to propose this new model. 
Section \ref{sec:3dPextraction} shows the fusion based model of \textit{3DPyraNet}. Section \ref{sec:RandD} gives details about the used benchmark datasets and achieved results. Finally, Section \ref{sec:Conc} concludes this paper. 
%
\section{\textit{PyraNet}}
\label{sec:2dpyranet}
\begin{figure}
\vspace{-2.8em} 
\centering
\resizebox{0.25\textwidth}{!}{
 \includegraphics{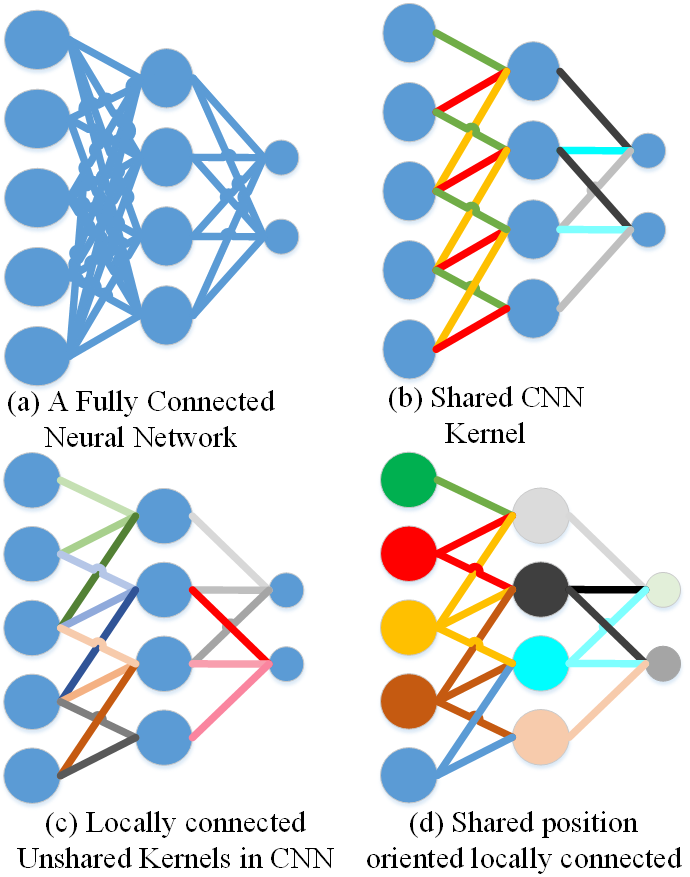}
}
\caption{Types of weighting schemes / receptive fields}
 \label{fig:weightSchemeVSOthers} 
\vspace{-2.0em}
\end{figure}

This section discusses some key features of the base model (\textit{PyraNet}) as well as some of its recent modifications. 
\textit{PyraNet} 
was an inspiration of the pyramidal \textit{NN} model reported in \cite{Cantoni2002} with 2D and 1D layers. 
The diversity with respect to original model \cite{Cantoni2002} is that the coefficients of a \textit(RF) were adaptive and it performed feature extraction and reduction in lower 2D layers. 
Finally, a 1D layer is adopted for classification of an image.
\newline
\textit{PyraNet} is composed of consecutive weighted sum pyramidal layers followed by FC layers. 
It is similar to \textit{CNN} \cite{Lecun1998} but without pooling layers. However, the two main differences between \textit{PyraNet} and \textit{CNN} are:
i) \textit{PyraNet} does not perform a convolution operation rather it performs weighted sum or correlation (\textit{WS} or \textit{CORR}) operation over the 2D RF ii) 
weight parameters are not in the form of a kernel that slides over the whole image, rather each output neuron has a unique local kernel specifically assigned to it. 
These temporary kernels are formed based on the position of input neurons in a \textit(RF) and their corresponding weights position on a 2D weight matrix having equal size as the input image/feature map size. This results in a unique locally connected (LC) kernel for each output neuron. 
Fig. \ref{fig:weightSchemeVSOthers} shows the difference between weight structure in a traditional FC NN (a), \textit{CNN} weight sharing (b), unshared LC weights (c), and the partially shared LC weights (d). The number of parameters in Fig. \ref{fig:weightSchemeVSOthers} (a), (b), (c) and (d) are 28 (20+8), 6 (3+3), 18 (12+6), and 9 (5+4), respectively. The base model of \textit{CNN} (Fig. \ref{fig:weightSchemeVSOthers} (b)) contains the least number of parameters compare to \textit{PyraNet} (Schema in Fig. \ref{fig:weightSchemeVSOthers} (d)), however,  usually \textit{CNN} have multiple kernels to produce multiple output maps (increase features), hence, as a result the number of parameters increases as compared to \textit{PyraNet}.   
\newline
The LC kernels in \textit{PyraNet} are partially shared with another neuron (based on overlap value) as shown with same color connection in Fig. \ref{fig:weightSchemeVSOthers} (d). 
Further, \textit{PyraNet} doesn't use any pooling layer for the reduction of dimensions, rather the dimensions are reduced by the stride/small-overlap of the kernel at each layer. 
During training, both \textit{CNN} and \textit{PyraNet} follow the same back-propagation (BP) technique for learning parameters with cross-entropy (\textit{CE}) loss function. However, due to the weighting scheme, the BP algorithm is updated in accordance with the new scheme (details can be found in \cite{Phung2007}).  
\textit{PyraNet} achieved 96.3\% accuracy, similar to \textit{SVM} for gender recognition and 5\% more than \textit{CNN} with same input size images of FERET dataset. 
\newline
Bruno et. al. \cite{Fernandes} introduced the concept of \textit{RF} inhibition in \textit{PyraNet} known as \textit{I-PyraNet}. In combination with the 2D-Gabor filter, it achieved good results for face detection on the MIT CBCL dataset. Later, it was used in \textit{SCRF} model \cite{fernandes2009nonclassical} for image segmentation and showed promising results.  
R. Uetz and S. Behnke \cite{Uetz2009}, presented an optimized, deeper and wider version of \textit{PyraNet}. In this model, feature maps were increased at each higher layer by using several kernels and adding reduced size maps of the previous layer to current layer. This model was evaluated on more difficult and bigger datasets i.e. NORB and LabelMe achieving 2.87\% and 16.27\% error, respectively. 
However, it didn't follow the pure pyramidal structure which results in huge cost in-terms of memory and time complexity.
\section{3DPyraNet}
\label{sec:3dpyranet}
The proposed 3D pyramidal architecture is based on the concept of coarse to fine refinement or the decision making pyramidal structure of a brain. This approach is widely used in \textit{NN} models \cite{Cantoni2002,Jose2013a,Chen2014,Wang2015,Long2015}. 
In addition, the structure of image pyramids and \textit{NN} is also quite similar. 
For this reason, \textit{3DPyraNet} model is developed by taking inspiration from an early pyramidal \textit{NN} model \cite{Cantoni2002} and image pyramid approach. 
A recent Deepface model \cite{Taigman} adopted unshared LC layers that has individual unshared kernel for each output neuron, resulting in enormous increase in the number of parameters. Same LC concept is adopted in \textit{CiC} model \cite{Pang2016} to reduce the number of parameters in the fully dense \textit{MLP} layer of the modified network in a network (\textit{NiN}) model \cite{NINLinCY13}. Parameters reduction not only reduces memory consumption and computational time but also increases it's generalization power \cite{Goodfellow2016Book, Pang2016}. Further, Y. Pang et. al, showed experimentally that unshared LC kernels perform better than shared kernels in a convolution operation for reduction of test error. The ability of shared kernels to reduce parameters and the impact of unshared LC kernels on the generalization power and reduction of test error motivated us to adopt a partially shared position oriented weight scheme as shown in Fig. \ref{fig:weightSchemeVSOthers} (d). 
Furthermore, due to the position oriented feature it capture the required spatial information as a whole for recognizing actions/dynamic scenes in the videos. 
\newline
To take advantage of temporal information in the videos, we adopted a 3D structure by taking motivation from 3D Convolutional Neural Networks (\textit{3D-ConvNet} and \textit{3DCNN}) models \cite{Baccouche11a,Ji2013a}. 
Model starts with a big input data stream, and then extracts sets of feature maps with randomly initialized several sets of weight matrices.
These feature maps are continuously refined at each higher layer until the model achieves a reduced most discriminative set of feature vector for the classification of targeted application area in the videos. Motivation behind this model was to reduce ambiguity in extracted features and eventually enhancing the performance. 
The objective is to highlight that giving pyramidal structure (despite having less feature maps and hidden layers) to a model can improve results as compare to non-pyramidal models.


\begin{figure*}
\centering
\resizebox{0.90\textwidth}{!}{
\includegraphics{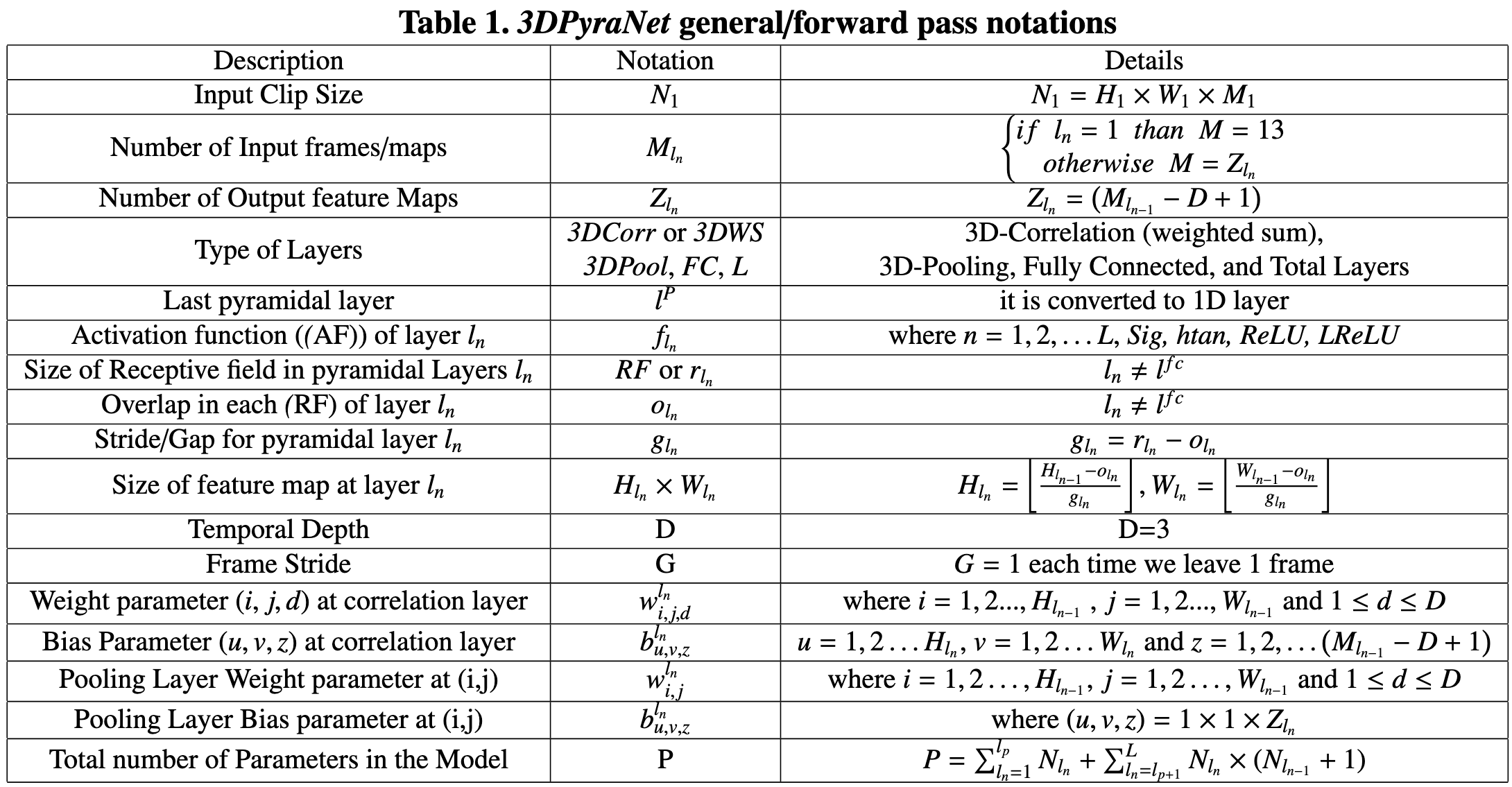}
 }
 \label{fig:table2}
 \end{figure*}

\subsection{Weighting Scheme}
\label{sec:WS}
\begin{figure*}
\centering
\resizebox{0.60\textwidth}{!}{
\includegraphics{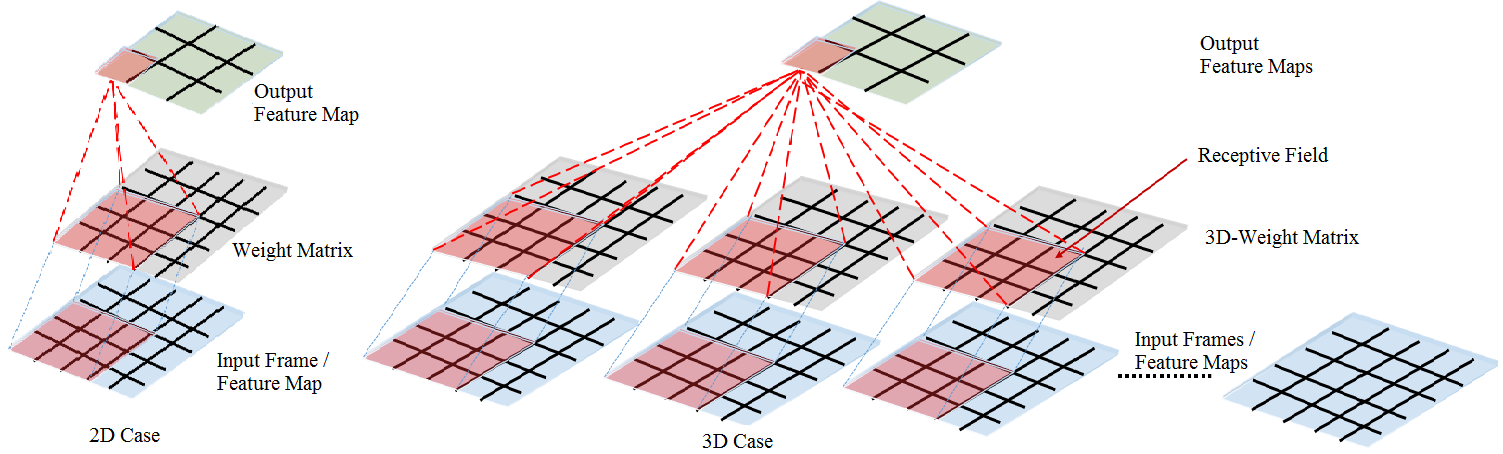}
 }
 \caption{Weighted scheme for 2D vs our 3D Scheme }
 \label{fig:3dWS}
 \end{figure*}

An important characteristic of popular convolutional deep models is their weight-sharing concept that gives an edge over other \textit{NN} models. This property reduces large number of learning parameters, however, also reduces the effectiveness of those fewer parameters. 
The weight scheme and BP technique in \textit{PyraNet} is adopted and modified for 3D structure. 
\newline
\textbf{Modified Weighting Scheme:} 
The concept of parameter sharing may not be so useful in some cases \cite{Pang2016}. As an example, if completely different features should be learned on one spatial position of the image than another, e.g. for face images that have been centered in the image, might need to learn various eye-specific or hair-specific or the relation of their features in different spatial locations.
 Similarly, in practical example of \textit{DSR} of a beach scenario, clouds or sky is always expected on upper position with sandy texture and water waves on the bottom. In such cases, it is better to avoid the traditional sharing scheme, and instead use a partial sharing scheme that may give additional power to the model.
\newline
In our 3D weighting scheme, three weight matrices are used at a time to incorporate temporal part. At the time of computation, each output neuron gets a unique 3D kernel from this 3D weight matrix as shown by 3D case in Figure. \ref{fig:3dWS}; 
The input frame and the weight matrix are of same size.
An output neuron is the sum of three weighted sum outputs of same \textit(RF) in three consecutive frames and the 3D weight matrix. It incorporates the temporal information from the given input frames. 
The weights are randomly initialized with respect to \textit(AF) and type of layer by taking care of the suggested techniques \cite{Glorot10understandingthe,orr2003neural,bengio2012practical}. 
\newline
\textit{RF} and \textit{O} are the two main tunable parameters for handling the performance of a network. 
$RF \times RF \times D$ is the mask used at a specific time that does the correlation operation between input and weight matrix. 
\textit{D}is the length along the temporal dimension. \textit{RF} also represented by \textit{r} being the height or width of a \textit{RF} kernel (i.e. $2, 3, 4,\dots$). \textit{O} can be any value less than \textit{RF}. 
Each weight parameter is shared locally in the \textit{RF} of few of the output neighboring neurons. Weight sharing in this 3D weight matrix approach is different than traditional sharing of \textit{CNN}. It is minimal and depends on overlap value, i.e. in worst case it can be just one time, otherwise depend on the overlap value. Even then it reduces a large amount of parameters as compared to recent DeepFace model \cite{Taigman}. 
\subsection{Proposed Architecture}
\begin{figure}
\centering
 \includegraphics[scale=0.3]{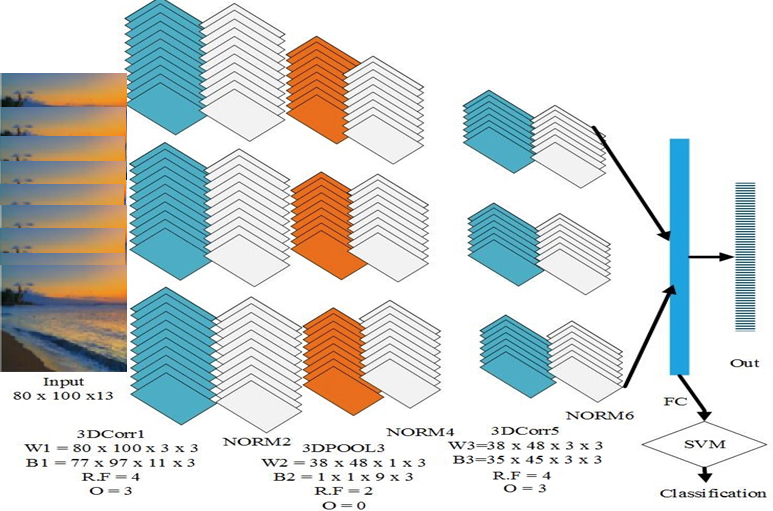}
\vspace{-1em}
 \caption{Proposed 3DPyraNet (other than SVM). With SVM it becomes 3DPyraNet-F. Blue, Gray, brown, and bright blue represents \textit{3DCORR}, normalization, pooling, and fully connected layer} 
 \label{fig:ourModel}
\vspace{-1.5em}
 \end{figure}

The basic \textit{3DPyraNet} model has three main hidden layers.
The given input is in binary/gray form with simple, unsophisticated pre-processing, unlike \cite{Ji2013a} deep model for \textit{AR} from videos. 
Table. \ref{table:architecture3DPyranet} describes some of the notation and their values used in forward propagation phase of\textit{3DPyraNet}. 
In general, the temporal part gives a correlation between the object/action/scene in consecutive frames of a video. The model starts with a \textit{3DCORR} layer as shown in Figure.~\ref{fig:ourModel}. \textit{3DCORR} extracts feature maps containing spatial as well as temporal information from the given input clip. \textit{3DPool} is introduced not only to reduce the resolution and computation time, but also to tackle translation invariance problem and avoiding over-fitting. 
The output from these correlation and pooling layers represents high-level features in the data. Essentially, these layers provide a meaningful, low-dimensional, and invariant feature space. This low-dimensional space is given to a \textit{FC} layer to learn a possibly non-linear function in that space. Finally, it classifies the given sample in its respective category.
\newline
\textit{3DPyraNet} and its extensions are shown in a single generalized model in Figure. \ref{fig:ourModel}. In this work, basic \textit{3DPyraNet} consists of mainly two \textit{3DCORR} layers, a \textit{3DPOOL}, and a \textit{FC} layer.
The extensions include a linear-SVM classifier layer which is discussed in section \ref{sec:3dPextraction}. 
\subsubsection{3D Correlation Layer}
\label{sec:3dcorrlayer}
An action is defined by the recognition of consecutive similar activity or pose of a human body over a continuous time span. 
Substantially, a dynamic scene can be recognized by similar structure, e.g. a beach scene might be characterized by drifting overhead clouds, mid-scene water waves and a foreground of static sandy texture. 
Therefore, CORR operation with proposed weighting scheme is most suitable for recognizing/learning similarity from actions/scenes in videos due to the existence of correlation in consecutive frames.
\newline
\textit{3DPyraNet} uses its 3D structure to incorporate the spatial as well as temporal information from the given input frames. 
The weight matrix generates sparse features as compared to a traditional convolutional kernel. 
Several sets of these 3D matrices are used to extract varied features. It generates multiple types of feature maps from the same given clip (set of frames).
The activated resulted map is normalized with simple zero mean and unit variance before passing it as input to the next layer. 
This normalization not only enhances accuracy by 4-5\%, but it also helps in faster convergence of the network. 
The \textit{3DPyraNet} perform \textit{3DCORR/3DWS} using a 3D kernel shown in Fig. \ref{fig:3dWS}. The output neuron $y_{u,v,z}^{l_{n}}$ on $z$ feature map in the $l_{n}$ layer is given by Eq.~\ref{eq:1}.  
\begin{equation}
\label{eq:1}
\resizebox{0.4\textwidth}{!}{$
y_{u,v,z}^{l_{n}} = f_{l_{n}} \left (\sum_{d=1}^{D} \sum_{(i,j,m) \in {R_{(u,v,z)}^{l_{n},d}}} \left ( \left ( w_{(i,j,d)}^{l_{n}}~ . ~  y_{(i,j,m)}^{l_{n-1} } \right ) + b_{(u,v,z)}^{l_{n} } \right ) \right )
$}
\end{equation}
Where $f_{l_{n}}$ represents an \textit(AF) used at current layer $l_{n}$. In these models, 'D' is $3$ as shown in Table. \ref{table:architecture3DPyranet}. The output neuron position is represented by $(u,v)$ at the current output feature map ($z$). This $z$ is generated by a set of input maps ($m$) in the temporal direction, where $m$ is calculated by '$d+z-1$' from layer $l_{n-1}$ as shown in Eq. \ref{eq:2} third row. 
\begin{equation}
\label{eq:2}
\resizebox{0.3\textwidth}{!}{$
R_{(u,v,z)}^{l_{n},d} =\left\{ \begin{matrix}
 ( i,j,m ) \mid {  ( u-1 ) + 1 \leq i \leq (u-1) + r_{l_{n}} };\\ 
~~{   (v-1) + 1 \leq j \leq (v-1) + r_{l_{n}} }; \\
~~{  (d_{low}+z-1) \leq m \leq (d_{high}+z-1)} 
 \end{matrix} \right\}
$}
\end{equation}
Here, $d_{low}$ and $d_{high}$ are $1$ and $D$, respectively, due to the size of the kernel minimum and maximum temporal depth. The set of neurons of a \textit{RF}, i.e. $i,j$ in the current map $m$ at the lower layer is calculated by Eq. \ref{eq:2}. Where $R_{(u,v,z)}^{l_{n},m}$ represents the \textit{RF} for each neuron ($u,v$) in $z$ output map. Here, $r_{l_{n}}$ in the Eq. \ref{eq:2}. represents the size of the \textit{RF} at layer $l_{n}$.
In case of biases, unlike \textit{CNN's}, \textit{3DPyraNet} does not use one bias for each output feature map, rather it uses one bias for each neuron in an output feature map. To have a pyramid structure in the model and to extract varied features from the input, we used three 3D weight matrix at first layer. 
Unlike \textit{3DCNN}, we did not increase set of kernels at each higher layer, rather kept it fixed. \textit{3DPyraNet} reduced feature maps by two at each upper layer.
\textit{RF} and \textit{O} are tuned for handling the performance. The \textit{AR} model uses \textit{RF} and \textit{O} size of 4 and 3 in \textit{3DCORR1} layer, respectively. Whereas, it is 3 and 2 in its \textit{3DCORR5} layer, respectively. Similarly, the model for \textit{DSR} uses \textit{RF} and \textit{O} size of 4 and 3 in both its \textit{3DCORR1} and \textit{3DCORR5} layer, respectively.  
\newline
To analyze what these weight matrices can learn, 
3D weight matrices are visualized for the first layer. Initially, feature maps produced from \textit{WS} kernels were sparse as compared to a convolutional kernel. After training the model, the maps as well as the weight matrices became similar to a smooth blurred image of the input sequences but different from each others in terms of texture, illumination, and the position of most activated neurons. 
\subsubsection{3D Temporal Pooling Layer}
\label{sec:poollayer}
The position oriented weight matrix approach has a slight deficiency of not learning, translation and scale invariant features. A 3D temporal max pooling layer (\textit{3DPOOL}) is introduced to overcome these limitations. It returns the maximum value among the three \textit{RF}s. This helps in removing non-maximum values that reduce computation for higher layers as well as provide translation invariance and robustness. Further, it helps in reducing the dimensionality not only in spatial domain but also in the temporal domain and maintaining the pyramidal structure of the model.   
\newline
In traditional pooling layers there are no weight parameters or bias's, \textit{3DPyraNet} model consists of a weight parameter for each output maximum value among the three referenced fields. Each maximum value is multiplied with a weight parameter and then a bias is added. Finally, this resultant signal is passed through an \textit(AF). The output max pooled value for neuron ${y}^{l_{n}}_{u,v,z}$ is calculated by Eq.~\ref{eq:pooling3d}. 
\begin{equation}  
\label{eq:pooling3d}
\resizebox{0.4\textwidth}{!}{$
{y}^{l_{n}}_{u,v,z}{=}{f}_{l_{n}}~\left(\left({w}^{l_{n}}_{u,v}~ . ~~ \max_{1 \leq d \leq D} \left(\max_{(i,j,m) \in ~{R}^{l_{n-1},d}_{u,v,z}} ~~ \left(y^{l_{n-1}}_{i,j,m}\right) \right) \right) + ~b^{l_{n}}_{u,v,z}\right)
$}
\end{equation} 
Here, ${R}^{l_{n},m}_{{u,v,z}}$ calculates the range for $(i,j,m)$ indices, i.e. $i_{low}$, $j_{low}$, $i_{high}$, $j_{high}$, $m_{low}$ and $m_{high}$ as being calculated by Eq.~\ref{eq:2} in previous layer. 
\textit{RF} and \textit{O} is taken as 2 and 0 in \textit{3DPOOL3} layer. 
\subsubsection{Fully Connected Layer}
\label{sec:FFFullyconnected}
Maps from \textit{3DPOOL3} layer are passed through the normalization (\textit{NORM4}) layer. Its output is processed with another \textit{(3DCORR5)} and \textit{(NORM6)} layer. The resultant normalized discriminative feature maps are converted into a 1D column vector that consists of motion information encoded in multiple adjacent frames. It is used as a FC layer for classification. It can be extended to multiple 1D \textit{FC} layers, depending on the complexity of the target application area. The size of this vector depends on the input size, total number of layers (\textit{L}), \textit{RF}, \textit{O}, and \textit{D}.
%
\begin{equation}
\label{eq:fullconOutput}
\resizebox{0.3\textwidth}{!}{$
y_{u,v,z}^{l_{n}} = f_{l_{n}} \left ( \sum_{i=1}^{I} \left ( \left ( w_{(i,v,z)}^{l_{n}} ~.~  y_{(i,1,1)}^{l_{n-1} } \right ) + b_{(1,v,1)}^{l_{n} } \right ) \right)
$}
\end{equation}
Here, $((i,d)_{low},(i,d)_{high})$ are $1$ due to $u$ and $z$ being 1. $l_{n}=L$ means that it is the final output layer otherwise, it is a 1D FC layer that is given as input to successive 1D layer until finally calculating the output layer. 
Weight update is done using conventionalBP algorithm with a stochastic gradient decent (\textit{SGD}) approach for minimizing the error. 
\subsection{3DPyraNet Training}\label{sec:3dpyranetBP}
A fast training algorithm must be devised to learn recognition task efficiently. 
\cite{Phung2007} suggests that \textit{CE} perform similar or better than mean squared error (\textit{MSE}). 
Therefore, the \textit{CE} error function is adopted that calculates the posterior probability membership for each action/scene class.
Delta rule given in Eq.~\ref{eq:deltarule} is used to update the weight parameters. 
\begin{equation}
\label{eq:deltarule}
\resizebox{0.2\textwidth}{!}{$
w^{l_{n},new}_{u,v,d}{ =\ }w^{l_{n},old}_{u,v,d}~~{-}~~ \varepsilon ~~~ \frac{\partial E}{\partial ~w^{l_{n}}_{u,v,d}}
$}
\end{equation}
Where $\varepsilon$ is the learning rate that controls the oscillation during training. 
In case of \textit{AR}, $\varepsilon$ is initiated with 0.00015 and reduces it by a factor of 10\% after each 10 epochs. 
Likewise for \textit{SR}, $\varepsilon$ is initiated with $0.000015$ and reduces it by a factor of 10\% after each 4 epochs. 
Batch size of 200 and 100 is used for \textit{AR} and \textit{SR}, respectively. 
\newline
The $\frac{\partial E}{\partial w^{l_{n}}_{u,v,d}}$ is calculated to update the weights at each layer. We divide it into two steps: 
first step calculates error sensitivity or local error for each neuron, while in the second step, weight gradients are calculated that update the learnable parameters. Calculating error gradients for FC layer is straight forward like a multi-layer perceptron. However, in pyramidal layers it becomes complicated. 
\begin{figure*}
\centering
\resizebox{0.90\textwidth}{!}{
\includegraphics{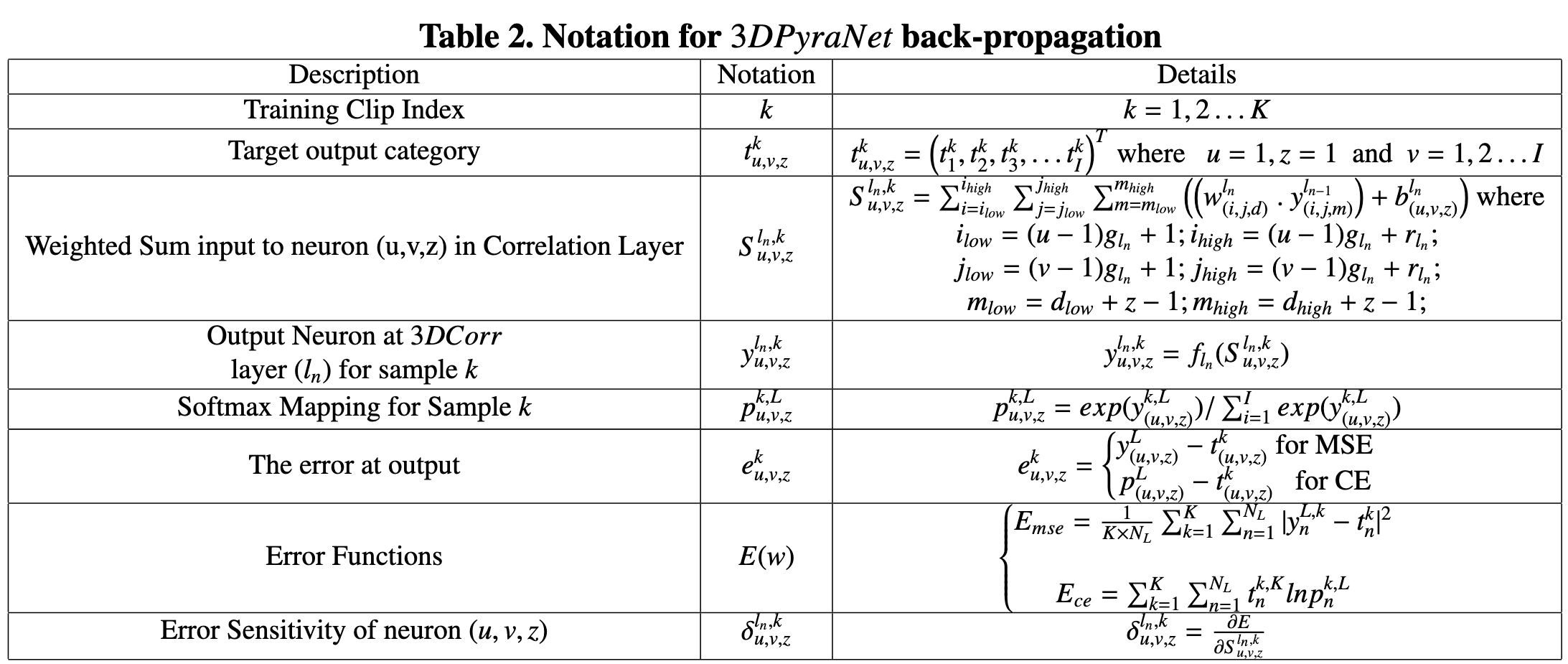}
 }
 \label{fig:table2}
 \end{figure*}

\subsubsection{Last Layer (L)}\label{sec:BPLastLayer}
The partial derivative of error with respect to the input is calculated for calculating the local gradient or error sensitivity ($\delta^{L}_{u,v,z}$) at the layer ($L$) with Eq. \ref{eq:BPOutLocalSensitivity}
\begin{equation}\label{eq:BPOutLocalSensitivity}
\delta^{L,k}_{u,v,z} = e^{k}_{u,v,z} ~f^{{'}}_L~ \left(S^{L,k}_{(u,v,z)}\right) 
\end{equation}
In case of MSE: $e^{k}_{u,v,z} = y^{L}_{(u,v,z)} - t^{k}_{(u,v,z)}$ \newline
In Case of CE: $e^{k}_{u,v,z} = p^{L}_{(u,v,z)} - t^{k}_{(u,v,z)}$ \newline
Where, $p^{L}_{(u,v,z)} = exp(y^{L}_{(u,v,z)}) / \sum_{i=1}^{I}exp(y^{L}_{(u,v,z)})$	\newline
The error ($e^{k}_{u,v,z}$) in case of \textit{MSE} represents the difference between network output and the target output. Whereas, in case of \textit{CE} it is the difference between posterior probability ($p^{L}_{(u,v,z)}$) and the target output. 
The difference arises in final layer for formulating the derivatives when we use different error functions. Otherwise, the rest of the equations in all layers remains the same while using any error function. 
Here, $(S^{L,k}_{(u,v,z)})$ represents the weighted sum for neuron $(u,v,z)$ and $f^{'}_L$ represents the inverse of an \textit(AF) at layer $L$. 
\subsubsection{Full Connected Layer (L-1)}\label{sec:BPFCLayer}
Eq. \ref{eq:BPOutLocalSensitivity} calculates the local error for the output neurons. Now to BP this error in 1D FC layers, we will calculate error sensitivity and then error gradients. However, this is not simple as compared to the output layer. Here, error is on output layer, that has to be transferred through connections to each neuron. 
$\delta^{l_{n},k}_{u,v,z}$ in equation \ref{eq:BPerrSens1DLayer} represents sensitivity of neuron $(u,v,z)$ at FC layer.
\begin{equation}
\resizebox{0.3\textwidth}{!}{$
\delta^{l_{n},k}_{u,v,z}{\rm =}~~f^{'}_{l_{n}}~\left(S^{l_{n},k}_{u,v,z}\right)\sum_{d=1}^{D} \sum^{j_{high}}_{j{=1}}{\delta^{l_{n+1},k}_{i,j,m} ~ w^{l_{n+1}}_{u,v,d}}
$}
\label{eq:BPerrSens1DLayer}
\end{equation}
Where $l_{n+1}$ is the upper layer (output) and $'l_{n}'$ is the 1D FC layer. We have used $i,j,m$ that are linked with current $u,v,z$. As it is a \textit{FC} layer, therefore, it has only one summation in the equation to change the variable $j$ (as other remains constant).
 In this case, $(i,m)$ are constants and are equal to 1 due to 1D vector.
 Similarly, $d_{high}$ is the number of neurons in the upper layer. Now actual weight gradients are calculated to update our weights.   \newline
\textbf{Weight Gradients at 1D or Fully Connected Layer:}
The weight gradients (Eq. \ref{eq:BPweightGradientFC1}) of 1D \textit{FC} layer are computed by the product of local gradients calculated in Eq. \ref{eq:BPerrSens1DLayer} and their respective inputs that generated the output in the forward propagation.
\begin{equation}
\label{eq:BPweightGradientFC1}
\resizebox{0.3\textwidth}{!}{$
\frac{\partial E}{\partial w^{l_n}_{i,j,d}} = ~\sum^K_{k=1} \sum^{j_{high}}_{j_{low}=1} {\delta^{l_{n},k}_{u,v,z}~y^{l_{n-1},k}_{i,j,m}} 
$}
\end{equation}
$\delta^{l_{n},k}_{u,v,z}$ represents the error sensitivity form the upper layer. As it is a 1D vector, therefore, $i_{high}$ and $m_{high}$ are equal to 1. Therefore, it will run for $j=j_{high}$, i.e. the number of neurons in the vector. 
Whereas, $K$ represents the total number of sample frames in a batch. This weight gradient is similar to calculating weight gradients of a \textit{FC} layer in a MLP. 
Eq. \ref{eq:BPerrSens1DLayer} and \ref{eq:BPweightGradientFC1} can be used for all the layers between output and last pyramidal layer ($l^{P}$), i.e. $l^{P}<l_{n}<L$. Rather, same equations are used for layer $l^{P}$. The only difference is that in case of $l^{P}$, after calculating error sensitivities and weight gradients it is rearranged in 3D structure. 
\newline
\textbf{Bias gradient for 1D or FC Layer:} 
The biases are updated with the same error sensitivities. However, $\frac{\partial E}{\partial b^{l_n}_{i,j}}$ is calculated by eq. \ref{eq:BP1DBiasGrad}. 
\begin{equation}
\label{eq:BP1DBiasGrad}
\resizebox{0.2\textwidth}{!}{$
\frac{\partial E}{\partial b^{l_n}_{i,j,d}} = ~\sum^K_{k=1} \sum^{v_{high}}_{v_{low}=1} {\delta^{l_{n},k}_{u,v,z}} 
$}
\end{equation}
In 1D case, $i$ and $d$ are equal to 1, only $j$ represents the number of output neurons for which their is one bias value. Therefore, the bias gradient is calculated by the summation of all the error sensitivities of that position in all the samples $K$. 
\subsubsection{3D Pyramidal Layer}\label{sec:BP3DWsCorrLayer}
After calculating the error gradients at \textit{FC} 1D layers, the error is BP to update the weight parameters at 3D pyramidal layers. 
Error sensitivity ($\delta^{l_{n}}_{u,v,z}$) at pyramidal layer is calculated by eq. \ref{eq:BPerrSen3dPyr}.
\begin{equation}
\label{eq:BPerrSen3dPyr}
\resizebox{0.35\textwidth}{!}{$
\delta^{l_{n}}_{u,v,z}=f^{'}_{l_{n}} \left(S^{{l_{n}},~k}_{u,v,z}\right) . \sum_{d=1}^{D} \sum^{i_{high}}_{i=i_{low}} \sum^{j_{high}}_{j=j_{low}} \sum^{m_{high}}_{m=m_{low}}  \delta^{{l_{n+1}},k}_{i,j,m} ~ w^{{l_{n+1}}}_{i,j,d} 
$}
\end{equation}
where $u= 1,2,\dots H_{l}$, and $v= 1,2,\dots W_{l}$. For each $z$, error sensitivity is computed by respective maps $m$ assigned by $d+z-1$ from layer $l_{n+1}$.
In Eq. \ref{eq:BPerrSen3dPyr}, $i_{low},i_{high},j_{low}$ and $j_{high}$ are calculated by Eq. \ref{eq:BP3DpyrRF1}, \ref{eq:BP3DpyrRF2}, \ref{eq:BP3DpyrRF3}, and \ref{eq:BP3DpyrRF4}, respectively. 
\begin{equation} 
\resizebox{0.1\textwidth}{!}{$
i_{low}{\rm =}~\left\lceil \frac{u{\rm -}r_{{l_{n+1}}}}{g_{{l_{n+1}}}}\right\rceil {\rm +1}
$}
\label{eq:BP3DpyrRF1}
\end{equation}
\begin{equation}
\resizebox{0.1\textwidth}{!}{$
i_{high}{\rm =}~\left\lfloor \frac{u{\rm -}{\rm 1}}{g_{{l_{n+1}}}}\right\rfloor {\rm +1}
$}
\label{eq:BP3DpyrRF2}
\end{equation}
\begin{equation}
\resizebox{0.1\textwidth}{!}{$
j_{low}{\rm =}~\left\lceil \frac{v{\rm -}r_{{l_{n+1}}}}{g_{{l_{n+1}}}}\right\rceil {\rm +1}
$}
\label{eq:BP3DpyrRF3}
\end{equation}
\begin{equation}
\resizebox{0.1\textwidth}{!}{$
j_{high}{\rm =}~\left\lfloor \frac{v{\rm -}{\rm 1}}{g_{{l_{n+1}}}}\right\rfloor {\rm +1}
$}
\label{eq:BP3DpyrRF4}
\end{equation}
\textbf{Weight Gradients for 3D Pyramidal Layers:}
The same steps are taken for pyramidal layers as it is being taken for 1D layer, i.e. calculate the weight gradients by taking the product of the sum of the local sensitivities at higher layer that were in contact with the current neuron at lower layer.  
\begin{equation}  
\label{eq:BP3DCorrWeightGrad}
\resizebox{0.45\textwidth}{!}{$
\frac{\partial E}{\partial ~w^{l_{n}}_{i,j,d}} = \sum^K_{k=1} \sum^{m_{high}}_{m=m_{low}} {\left( y^{{l_{n-1}},k}_{i,j,m}~ \times ~ \sum^{u_{high}}_{u=u_{low}} \sum^{v_{high}}_{v=v_{low}} \sum^{z_{high}}_{z=z_{low}} \delta^{{l_{n}},k}_{u,v,z}\right)}
 $}
\end{equation}
here $i = 1,2,\dots H_{{l_{n}}}$, and $j = 1,2,\dots W_{{l_{n}}}$. Whereas, $m_{low}$ and $m_{high}$ are in the range such that: if $d=1$ than $m_{low}=d$ and $m_{high}=maps-2$, if $d=2$ than $m_{low}=d$ and $m_{high}=maps-1$, and finally if $d=3$ than $m_{low}=d$ and $m_{high}=maps$. '$maps$' represents total number of frames in that layer. However, $z_{low}$ and $z_{high}$ can be computed by Eq. \ref{eq:gradRF5} and \ref{eq:gradRF6}. 
\begin{equation}
\label{eq:gradRF1}
\resizebox{0.1\textwidth}{!}{$
u_{low}{\rm =}~\left\lceil \frac{i{\rm -}r_{l_{n}}}{g_{l_{n}}}\right\rceil {\rm +1}
$}
\end{equation}
\begin{equation}
\label{eq:gradRF2}
\resizebox{0.1\textwidth}{!}{$
u_{high}{\rm =}~\left\lfloor \frac{i{\rm -}{\rm 1}}{g_{l_{n}}}\right\rfloor {\rm +1}
$} 
\end{equation}
\begin{equation}
\label{eq:gradRF3}
\resizebox{0.1\textwidth}{!}{$
v_{low}{\rm =}~\left\lceil \frac{j{\rm -}r_{l_{n}}}{g_{l_{n}}}\right\rceil {\rm +1}
$} 
\end{equation}
\begin{equation}
\label{eq:gradRF4}
\resizebox{0.1\textwidth}{!}{$
v_{high}{\rm =}~\left\lfloor \frac{j{\rm -}{\rm 1}}{g_{l_{n}}}\right\rfloor {\rm +1}
$}
\end{equation}
\begin{equation}
\label{eq:gradRF5}
\resizebox{0.1\textwidth}{!}{$
z_{low}{\rm =}~\left\lceil \frac{m-D}{G}\right\rceil {\rm +1}
$}
\end{equation}
\begin{equation}
\label{eq:gradRF6}
\resizebox{0.1\textwidth}{!}{$
z_{high}{\rm =}~\left\lfloor \frac{m{\rm -}{\rm 1}}{G}\right\rfloor {\rm +1}
$}
\end{equation}
'G' represents the number of frames that are left after each 3DCORR. In our experiments, it is kept as one e.g. for first feature map, the input frames/feature maps are taken as 1,2, and 3, whereas for the second output map they are 2,3, and 4. 
\newline
\textbf{Bias gradient for 3D Pyramidal Layer:} 
The bias's are also updated with the same error sensitivities. However, $\frac{\partial E}{\partial w^{l_n}_{i,j,m}}$ is calculated by eq. \ref{eq:BP3DPyramidalBiasGrad}.
\begin{equation}
\label{eq:BP3DPyramidalBiasGrad}
\resizebox{0.18\textwidth}{!}{$
\frac{\partial E}{\partial w^{l_n}_{i,j,m}} = \sum^K_{k=1} ~ {{\delta^{{l_n},k}_{u,v,z}}}
$}
\end{equation}
The bias error gradient is the sum of all the error sensitivities of that position in all the maps from all the samples $K$. $b^{l_n}_{i,j,m}$ is the bias for neuron $(i,j)$ in map $m$. As we have one bias for each output neuron therefore $i = u$, $j = v$, and $m = z$.  
%
\subsubsection{Backward Temporal Pooling Layer}
\label{sec:BPtpooliingLayer}
The technique to calculate weight gradients at pooling layer is the same as BP in correlation layer. However, the difference arise due to error that BP only through the selected maximum neuron (in case of max pooling) among the three \textit{RF}s used in calculating the weight gradient. Eq. \ref{eq:BPpoolsen} calculates error sensitivity at pooling layers.
\begin{equation}
\label{eq:BPpoolsen}
\resizebox{0.4\textwidth}{!}{$
\delta_{u,v,z}^{l_{n},k}=\sum_{d=1}^{D} f_{l_{n}}^{'} \left(S_{u',v',z'}^{l_{n},k}\right) \sum^{i_{high}}_{i=i_{low}} \sum^{j_{high}}_{j=j_{low}} \sum^{m_{high}}_{m=m_{low}} \delta_{i,j,m}^{{l_{n+1}},k} . w_{i,j,d}^{l_{n+1}}
$}
\end{equation}
Where the indices $(u',v',z')$ represent the points when it attains the largest value in the \textit{RF} $R_{u,v,z}^{l_{n},d}$. 
\begin{equation}
\argmax_{u',v',z'} S_{u',v',z'}^{l_{n},k} := \{(u',v',z') | \forall (u,v,z) : S_{u,v,z}^{l_{n},k} < S_{u',v',z'}^{l_{n},k}\}
\end{equation}
It represents the maximum of the maximum values among the three \textit{RF}s calculated in the same manner as being done in selecting the max value in the forward propagation by Eq. \ref{eq:2}. $S_{u',v',z'}^{l_{n},k}$ is the weighted sum value resulted from the weight parameter $(w_{u,v})$ and the maximum value. The rest of the range, i.e. $i_{low},i_{high},j_{low}$ and $j_{high}$ are calculated by Eq. \ref{eq:BP3DpyrRF1}, \ref{eq:BP3DpyrRF2}, \ref{eq:BP3DpyrRF3}, and \ref{eq:BP3DpyrRF4}, respectively. 
For each $z$, error sensitivity is calculated by respective maps $m$, computed in the range of $m_{low}=z$ and $m_{high}=z+D+1$ from layer $l_{n+1}$. 
The ranges for \textit{RF}s are calculated by Eq. \ref{eq:gradRF1}, \ref{eq:gradRF2}, \ref{eq:gradRF3}, \ref{eq:gradRF4}, \ref{eq:gradRF5}, and \ref{eq:gradRF6}.
\newline
\textbf{Weight gradient for 3D pooling layer:} 
The weight gradient for 3D pooling layer is calculated as
\begin{equation}
\label{eq:BP3DPoolWeightGrad}
\resizebox{0.4\textwidth}{!}{$
\frac{\partial E}{\partial w^{l_n}_{i,j}} = \sum^K_{k=1}~ \sum^{M_{l_{n}}}_{m=1} ~ \left({y}^{l_{n-1},k}_{i,j,m} \right)  ~.~ \sum^{u_{high}}_{u=u_{low}} \sum^{v_{high}}_{v=v_{low}} \sum^{z_{high}}_{z=z_{low}} {{\delta^{{l_n},k}_{u,v,z}}}
$}
\end{equation}
Where $\frac{\partial E}{\partial w^{l_n}_{i,j}}$ calculates the weight gradients to be used in Eq. \ref{eq:deltarule} for updating the weight parameters at pooling layer, and $\left({y}^{l_{n-1},k}_{i,j,m} \right)$ is the maximum value selected in Eq. \ref{eq:pooling3d}. 
Range values ($v_{low}, v_{high}, u_{low}$, and $u_{high}$) are calculated by Eq. \ref{eq:gradRF1}, \ref{eq:gradRF2}, \ref{eq:gradRF3} and \ref{eq:gradRF4}, respectively. Eq. \ref{eq:gradRF5} and \ref{eq:gradRF6} are used for selecting corresponding maps, i.e. $z_{low}$ and $z_{high}$ containing error sensitivities. 
\newline
\textbf{Bias gradient for 3D pooling layer:} 
The biases are also updated with the same error sensitivities. However, $\frac{\partial E}{\partial b^{l_n}_{i,j}}$ is calculated by Eq. \ref{eq:BP3DPoolBiasGrad}. 
\begin{equation}
\label{eq:BP3DPoolBiasGrad}
\resizebox{0.3\textwidth}{!}{$
\frac{\partial E}{\partial b^{l_n}_{i,j,m}} = \sum^K_{k=1} ~ \sum^{u_{high}}_{u=1} \sum^{v_{high}}_{v=1} {{\delta^{{l_n},k}_{u,v,z}}}
$}
\end{equation}
The bias gradient calculation for pooling layer is similar to pyramidal layers. The difference is due to the reason that \textit{3DPyraNet} have only one bias for each output map in pooling layer. Due to which $i=1$ and $j=1$ in equation. \ref{eq:BP3DPoolBiasGrad}. Therefore, the error gradient in case of biases is the sum of all the error sensitivities of those maps for all the samples $K$. $b^{l_n}_{i,j,m}$ is the bias for all the neurons in that map $m$. $u_{high}$ and $v_{high}$ represents the total rows and columns in the feature map. Also, as \textit{3DPyraNet} has only one bias for each map, therefore, in this case $m$ is the same as $z$. 
\section{Features Fusion for Spatio-temporal Feature Learning}
\label{sec:3dPextraction}
\textit{3DPyraNet} generates sparse features as compared to convolutional kernel and are learned using modified BP with mini-batch SGD approach. 
A variety of deep architectures can be designed from \textit{3DPyraNet} based on its application, input image size, complexity, number of layers, or combination of multiple models to enhance the performance. However, here mainly two types of models are presented based on feature fusion, i.e. local and global fusion an inspiration from the work done in \cite{Karpathy2014}.
\subsection{3DPyraNet-F}
\label{model:F}
Selecting an optimal architecture is a challenging problem, since it depends on the specific application. 
A generalized model is shown in Fig.\ref{fig:ourModel} due to limited space. Mainly, it consists of two \textit{3DCORR} layers, a \textit{3DPOOL}, a \textit{FC} layer, and a linear-SVM classifier layer. 
Once convergence is achieved, features from the last \textit{Norm6} layer are extracted and fused in a single column feature vector. 
We call this a global/early fusion based model (\textit{3DPyraNet-F}) which is a balanced mix between the spatial and temporal information. These are incorporated in such a way that global information in both spatial and temporal dimensions are progressively accessed and provided to an \textit{SVM}. 
\textit{SVM} trains over these fused features. Finally, the trained \textit{SVM} model is used to classify the feature vectors extracted from the testing set using \textit{3DPyraNet}. One-vs-all criteria is used for classification.
The depth, width, and the resulting size of the feature vector in this network depends on the input size, \textit{RF}, and \textit{O} parameters at each layer. 
\subsection{3DPyraNet-F\_M}
\label{model:FM}
The difference between \textit{3DPyraNet-F} and \textit{3DPyraNet-F\_M} is in the fusion and construction of feature vectors. 
After \textit{3DPyraNet} converges, rather than just fusing all the features in one column vector, 
this model first locally fuse feature maps of the same set in one single column vector.  
Then, the resultant feature vectors are summed together and divided by the number of weight sets to derive their mean vector. 
This results in a smaller feature vector compared to the previous model as well as in faster processing. These features have a local impact due to their addition with other features maps. The rest of the model and network architecture is similar to \textit{3DPyraNet-F}. 
\par
Table. \ref{table:NetworkSt} shows the three models that we have used in our experiments. Two models for AR datasets whereas, the third model is for \textit{DSR}. There are three main differences among the three models. The first two are based on local and global fusion of features whereas the third is because of the input size given to the network. This difference exists due to different size input images of \textit{AR} and \textit{DSR} datasets. 
\begin{table}[t]
\vspace{-1em}
\caption{Feature map size and output classes at main layers in each model}
\label{table:NetworkSt}
  \centering
\resizebox{0.45\textwidth}{!}{
 \begin{tabular}{|c|c|c|c|c|c|} 
 \hline
 Model & \textit{3DCORR} & \textit{3DPOOL} & \textit{3DCORR} & \textit{FC} & Output \\ 
\hline
\textit{3DPyraNet-F} & $61 \times 45 \times 11 \times  3$ & $30 \times 22 \times 9 \times 3$ & $27\times 19\times 7\times 3$ & $10773$ & $6/10$ \\
\hline
\textit{3DPyraNet-F\_M} & $61\times 45\times 11\times 3$ & $30\times 22\times 9\times 3$ & $27\times 19\times 7\times 3$ & $3591$ & $6/10$ \\
\hline
\textit{3DPyraNet-F} & $77\times 97\times 11\times 3$ & $38\times 48\times 9\times 3$ & $35\times 45\times 7\times 3$ & $33075$ & $13/14$ \\  
\hline
\end{tabular}
}
\vspace{-2.0em}
\end{table}
\section{Results \& Discussion}
\label{sec:RandD}
Firstly, \textit{3DPyraNet} has been evaluated for \textit{AR} on the \textit{Weizmann}
and \textit{KTH}
datasets \cite{Schuldt2004,Blank2005}. 
Later, we show the enhancement in performance for \textit{AR} by \textit{3DPyraNet-F} and \textit{3DPyraNet-F\_M} over \textit{3DPyraNet}.
Further, \textit{3DPyraNet} is compared with state-of-the-art handcrafted feature descriptors as well as feature learning approaches.  
Secondly, we examined \textit{DSR} on the \textit{YUPENN} and \textit{MaryLand} datasets. Beside accuracy, another key advantage of \textit{3DPyraNet} model is discussed, i.e. having fewer trainable parameters. 
\newline
\textbf{Training}: Each model is trained on its respective dataset. Table. \ref{table:NetworkSt} shows a feature map size in the form of $w \times h \times m \times s$. Where 'w', 'h', 'm' and 's' represents width, height, number of maps, and weight sets, respectively. \textit{KTH} \& \textit{Weizmann} have similar input size i.e. $64\times48\times13$ with leave one frame out. Whereas, \textit{YUPENN} \& \textit{MaryLand} have $80\times100\times13$ with an overlap of 7 images for each clip. Training is done by SGD with mini batch size of 200 and 100 clips for \textit{AR} and \textit{SR} datasets, respectively. 
\par
In case of \textit{AR}, we start with a small learning rate, i.e. 0.00015 and then decrease it after every 10 epochs by multiplying it by 0.9. Whereas, in case of \textit{DSR} learning rate is taken as 0.000015. The learning rate is decayed after every 4 epochs by multiplying it with 0.9. Early stopping criteria is adopted where the training stops when the testing/validation accuracy stops improving. The \textit{RF} and \textit{O} sizes are shown in Figure. \ref{fig:ourModel} as well mentioned in section \ref{model:F}. A linear-\textit{SVM} trained with Sequential Minimal Optimization is used to analyze discriminative power of the learned features for recognition of the action/scene. %
\subsection{Datasets}
\label{sec:datasets}
\begin{figure}
\centering
 \includegraphics[scale=0.8]{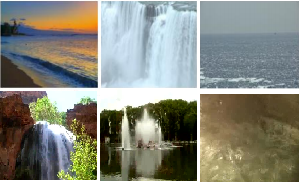}
\vspace{-1em}
 \caption{Samples from YUPENN ($1^{st}$ row) and MaryLand ($2^{nd}$ row) datasets}
 \label{fig:yupenMaryland}
\vspace{-2.0em}
 \end{figure}

\textit{Weizmann} and \textit{KTH} are well-known \textit{AR} datasets with actions such as Walking, Running, Jumping, etc. In weizmann and KTH, each action is done by 9 and 25 actors, respectively. These results in fewer videos per category hence increasing the complexity for a deep learning model.
In both datasets, \textit{3DPyraNet} uses an input sequence of $64\times48\times13$ consecutive frames but leaving one in the middle. With this scheme, there must be at least 25 consecutive frames that have a proper bounding box of the region of interest. 
3DSOBS \cite{Maddalena201465} is used to extract a person from the clip. 
\newline
\begin{figure}
\centering
 \includegraphics[scale=0.65]{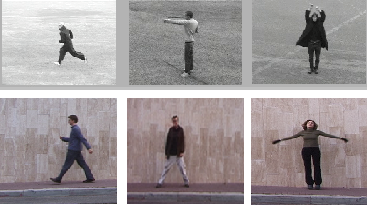}
 \vspace{-1em}
\caption{Samples from KTH ($1^{st}$ row) and Weizmann ($2^{nd}$ row) datasets}
 \label{fig:kthWizmn}
\vspace{-1.0em}
 \end{figure}
\textit{DS}s are categorized by a collection of dynamic patterns and their spatial layout, as recorded in small video clips. 
For example, a beach scene might be characterized by water waves and at its front a static sandy texture as shown in Fig. \ref{fig:yupenMaryland}.
These scenes are recorded by either static or moving cameras; thus, while scene motion is characteristic, it is not exclusive of camera induced motion. Indeed, \textit{DSR} from moving cameras has proven to be more challenging as compared to static cameras.
\textit{YUPENN} (static camera) consists of 420 videos (fixed size) of 14 scene categories (listed in Table.\ref{table:perClassComparison}).
Similarly, \textit{MaryLand} (non-static camera) consists of 130 (non-fixed size) videos of 13 scene categories (listed in Table. \ref{table:perClassCompa}).
For \textit{DSR}, we simply convert the RGB images in gray level images. 
These two datasets are tested only through \textit{3DPyraNet} and \textit{3DPyraNet-F} models.
%
\subsection{3DPyraNet}
In the first case, a network with two \textit{3DCORR} and a \textit{FC} layer is used to train and classify ten classes. The output of each \textit{3DCORR} layer is passed through a sigmoid or hyperbolic tangent function and then normalized throughout the network learning. Initial learning is not smooth and \textit{3DPyraNet} took around 450 epochs to converge. It provides accuracy of $80\%$ on the training set and $70\%$ on the testing set. 
As in most deep models, pooling plays an important role by providing translation invariance as well as reducing the dimensions. 
In addition, for faster convergence, avoidance of local minima, and improvement in performance, an extension of the rectified linear unit known as leaky rectified linear units ($LReLu$) \cite{maas2013rectifier} is utilized. This $LReLu$ in contrast to $ReLu$ allows a small non-zero gradient when the neuron is less than or equal to zero. This property overcomes the limitation of $ReLU$ and updates the weights even if stuck within zeros.
\newline
In second case, we used \textit{3DPOOL} and $LReLu$ beside \textit{3DCORR} layers.  
This resulted in high accuracy, i.e. 87\% (training) and 80.5\% (testing) and faster convergence i.e. within 200 epochs. Moreover, learning behavior during training was quite smooth compared to the previous model. 
\newline
\textit{3DPyraNet} is compared with deeper models having 5-8 hidden layers. To better evaluate \textit{3DPyraNet}, the mean accuracy is reported on five splits of training and testing datasets selected from the same Weizmann database as adopted for evaluation of several other models e.g. \textit{3D-ConvNet} model \cite{Baccouche11a}. To cross validate the results, the data is randomized in the same proportion by keeping in mind that equal number of sequences should exist for the small number of sequences e.g. 'skip' or 'running'. \textit{3DPyraNet} achieved $90.9\%$ accuracy for considering all ten classes in the dataset.
However, videos containing action 'skip' were brief. Many authors in literature did not use this category in their experiments. 
If we neglect the skip category, accuracy increases to $92.46\%$ as shown by \textit{3DPyraNet}(all-1) model in Table. \ref{table:KTHWEIZMANCOMPARISON}. However, for the rest of the categories, \textit{3DPyraNet} shows optimal results.
The performance on Weizmann is comparable with \textit{3D-ConvNet} model \cite{Baccouche11a}, which is impressive considering fewer number of hidden layers. Further, it overcame reported best result by \textit{3DConvNet} model, i.e. 88.26\% with an average of 91.07\% from ten tests using the same dataset and number of consecutive input frames.
\newline
In case of \textit{KTH}, a similar criteria to \cite{Baccouche11a} is used that took 9 out of 25 person's videos for testing. 
It should be noted that we faced the same problem in running videos i.e. having fewer frames than the minimum requirement of 13 due to fast movement of the person or camera zooming scenarios. We achieved $72\%$ accuracy over six classes and $74.23\%$ when 'running' is not considered.
\textit{3DPyraNet} and \textit{3DPyraNet}(all-1) in Table. \ref{table:KTHWEIZMANCOMPARISON} shows a comparison of the proposed model with the state-of-the-art models reported in literature.
\begin{table*}[t]
\begin{center}
\vspace{-1em}
\caption{Per class accuracies for MaryLand-in-the-Wild dataset}
\label{table:perClassCompa}
\resizebox{\textwidth}{!}{
 \begin{tabular}{|c|c|c|c|c|c|c|c|c|c|c|c|c|c|c|}
\hline
Model&Avalanche&Boiling Water&Chaotic Traffic&Forest Fire&Fountain&Iceberg Collapse&Landslide&Smooth Traffic&Tornado&Volcanic Eruption&Waterfall&Waves&Whirlpool&Overall\\
\hline
\begin{tabular}{@{}c@{}}C3D \cite{Tran2014a},\cite{ Feichtenhofer2016} \end{tabular} &90&90&90&80&60&60&70&80&80&90&40&100&80&78\\ 
\hline
\begin{tabular}{@{}c@{}}DPCF \cite{Feichtenhofer2016} \end{tabular} &90&60&100&90&80&50&80&70&80&90&70&100&80&80\\
\hline
\textit{3DPyraNet-F} &93&97&97&90&99&97&96&98&91&94&98&97&85&\textbf{95}\\
\hline
\end{tabular} 
}
\end{center}
\vspace{-1.0em}
\end{table*} 
%
\begin{table*}[t]
\begin{center}
\vspace{-1.5em}
\caption{Per class accuracies for \textit{YUPENN}}
\label{table:perClassComparison}
\resizebox{\textwidth}{!}{
 \begin{tabular}{| c | c |c|c|c | c |c|c|c | c |c|c|c | c |c|c|}
\hline
Model&Beach&Elevator&Forest Fire&Fountain&Highway&Lighting Storm&Ocean&Railway&Rushing River&Sky-Clouds&Snowing&Street&Waterfall&Windmill farm&Overall \\
\hline
\begin{tabular}{@{}c@{}}C3D \cite{Tran2014a,Feichtenhofer2016} \end{tabular} &97&100&100&83&97&93&100&97&100&97&97&100&97&100&97\\ 
\hline
\begin{tabular}{@{}c@{}}DPCF \cite{Feichtenhofer2016} \end{tabular} &100&100&97&93&100&100&100&100&100&100&97&100&97&100&\textbf{99}\\
\hline
\textit{3DPyraNet-F} &92&94&94&93&93&99&98&93&100&86&93&93&93&90&94\\
\hline
\end{tabular}
}
\end{center}
\vspace{-1.5em}
\end{table*} 
%
On the other hand, for \textit{KTH} dataset, \textit{3DCNN} \cite{Ji2013a} shown in Table. \ref{table:KTHWEIZMANCOMPARISON} used ROI's sequences extracted and classified by another \textit{CNN} based tracker. Whereas, M. Baccouche et al. for \textit{3DConvNet} model used simple raw input images but uses a  deep model as well as divided the KTH dataset into two sub datasets i.e. KTH1 and KTH2 \cite{Baccouche11a}. It was based on the complexity of the video sequences. They separated complex videos with multiple appearance of a person from the single appearance of a person in the videos.  
The model in \cite{Baccouche11a} report the results on a separate set with voting scheme. However, results on full KTH were not shown. 
We used random set of samples from full \textit{KTH} datasets. Only background subtraction is being done for extracting the binary ROIs containing human mask. This may contain half, not centered or unaligned ROIs as input. These unaligned ROIs can greatly affect the learning process and may have a high impact in reducing the classification rate. 
\newline
\textit{3DPyraNet} does not show optimal results as provided by \textit{3DConvNet} \cite{Baccouche11a} and \textit{3DCNN} \cite{Ji2013a}, but despite fewer layers it shows comparable results to some of the complex models as shown in Table.\ref{table:KTHWEIZMANCOMPARISON}. One of the most plausible reason is that deep models need more data to have a better understanding of their respective problems. Moreover, one of the reasons for our lower performance in case of \textit{KTH} dataset in comparison to \textit{3DConvNet} could be the reason that they divided (as previously mentioned) the complexity of the input data. 
Whereas, we tested our model on samples from the whole dataset.  
\newline
In case of \textit{DSR}, \textit{3DPyraNet} shows good performance for \textit{MaryLand} dataset despite the camera induced motion. In multi-class problem, Table. \ref{table:KTHWEIZMANCOMPARISON} shows that \textit{3DPyraNet} gave almost similar accuracy as \cite{feichtenhofer2013spacetime} with only $0.7\%$ difference. Currently, several deep models didn't report or evaluated their model on \textit{DSR} for multi-class problem, therefore \textit{3DPyraNet} can not be directly compared with them. In case of \textit{YUPENN} dataset, \textit{3DPyraNet} didn't perform as expected. One of the reasons could be that \textit{3DPyraNet} performs better when there is presence of motion in the videos as it is the case in \textit{MaryLand} dataset. 
Another reason could be that the model need further tuning to give optimal results in case of \textit{YUPENN} dataset.
\subsection{3DPyraNet-F \& 3DPyraNet-F\_M}
\label{sec:3dPasFE}
\textit{3DPyraNet-F \& 3DPyraNet-F\_M} are first evaluated for \textit{AR} with \textit{Weizmann} and \textit{KTH}. 
Its enhancement over \textit{3DPyraNet} is shown. It is compared with state-of-the-art handcrafted feature descriptors as well as feature learners.  
Secondly, \textit{3DPyraNet-F} is examined for \textit{DSR} with the help of \textit{YUPENN} and \textit{MaryLand}. \textit{3DPyraNet-F\_M} shows poor result for \textit{DSR} due to huge reduction in features. Therefore, it not discussed in detail. In the end, a key advantage of the proposed model is being discussed, i.e. its fewer trainable parameters. Training is done in the same way as being done for simple \textit{3DPyraNet}. 
%
\begin{table*}[!h]
\begin{center}
\vspace{-0.5em}
\caption{Accuracies for Action and Dynamic Scene datasets, Layers represents main layers, Parameters are in million, and size is in MB}
\label{table:KTHWEIZMANCOMPARISON}
\resizebox{\textwidth}{!}{
 \begin{tabular}{|c|c|c|c|c|c|c|}
\hline
 \textbf{Model(classifier)}&\textbf{Weizmann}&\textbf{KTH}&\textbf{YUPENN}&\textbf{MaryLand}&\textbf{Layers}& \begin{tabular}{@{}c@{}} \textbf{Parameters in Millions (Size in MB)} \end{tabular} \\
\hline\hline
 \begin{tabular}{@{}c@{}}  3D-ConvNet \cite{Baccouche11a}\end{tabular} & 88.26 & 89.40 &-&-&7&0.01717 (0.31) \\ [0.5ex] 
\hline
 3DCNN \cite{Ji2013a} & - & 90.2 &-&-& 6 &0.00511(0.09) \\ [0.5ex] 
\hline
 Cuboids \cite{Wang2009} & - & 90 & - &-& - &-\\ 
\hline
 \begin{tabular}{@{}c@{}}Gabor3D+HOG3D (SVM) \cite{Maninis2014} \end{tabular} & - & 93.5 & - &-& -&- \\ 
\hline
 \begin{tabular}{@{}c@{}}3DSIFT (SVM) \cite{Scovanner2007} \end{tabular} & 82.6 & - & - &-& - &-\\  
\hline
 \begin{tabular}{@{}c@{}}HOG+HOF+MBH+Trajectories(SVM) \cite{Wang2011} \end{tabular} & - & 94.2 & - &-& -& - \\ 
\hline
 C3D (SVM) \cite{Tran2014a} & - & - & 98.1 & 87.7 & 15 & 17.5 (305.14)\\ 
\hline
 ImageNet \cite{Tran2014a} & - & - & 96.7 & 87.7 & 8 & 17.5 (305.14)\\ 
\hline 
 Schuldt (SVM) \cite{Schuldt2004} & - & 71.7 &-&-& -& - \\
 \hline
 Dollar (SVM) \cite{Dollar2005} & - & 81.2 &-&-& - & -\\
\hline
\begin{tabular}{@{}c@{}} 3DHOG+Local weighted SVM \cite{Weinland2010} \end{tabular} & 100 & 92.4 & - &-& - & -\\
\hline
3DPyraNet & 90.9 & 72 & 45 & 67 & 4 & 0.83 (14.58)\\
 \hline
\textbf{3DPyraNet-F} & 98.99 & 93.42 &93.67&\textbf{94.83}& 4 & 0.83 (14.58)\\  
\hline
\textbf{3DPyraNet-F\_M} & \textbf{99.13} & \textbf{94.083} &-&-& 4 & 0.83 (14.58)\\  
\hline
 \begin{tabular}{@{}c@{}}  Christoph's (SVM) \cite{Feichtenhofer2016} \end{tabular} &-&-& \textbf{99} & 80 & - & -\\
\hline
\begin{tabular}{@{}c@{}}  Christoph's (SVM) \cite{feichtenhofer2014bags} \end{tabular} &-&-& 96.2 &77.7& - & -\\
\hline
\begin{tabular}{@{}c@{}}  Theriault's (SVM) \cite{6619180} \end{tabular} & - & - & 85.0 & 74.6& - & -\\
\hline
\end{tabular}
}
\end{center}
\vspace{-2.5em}
\end{table*} 
\vspace{-0.5em}
\subsubsection{Action Recognition}
\textit{KTH} and \textit{Weizmann} are easy for an in-depth study due to less data and more classes, however, challenging as well due to less data for training a deep model. 
Features form the \textit{Norm6} layer of our trained model are extracted, fused, and fed to a linear-\textit{SVM} classifier. It classifies each class similar to what is done in  \cite{Tran2014a, Schuldt2004, Dollar2005}. However, we perform two types of extraction, i.e. local and global fusion of features as in \cite{Karpathy2014}.
In first case, \textit{3DPyraNet-F} feature vector becomes 10733 whereas, in the second case \textit{3DPyraNet-F\_M} feature vector consists of 3591 features. 
\textit{3DPyraNet-F} achieved a mean accuracy of 93.42\% in one-vs-all scenario. Further, (\textit{3DPyraNet-F\_M}) enhances the overall performance by $0.67\%$. Similarly to \cite{Schuldt2004, Ji2013a, Dollar2005}, despite fewer training examples, global fusion (\textit{3DPyraNet-F}) achieved optimal accuracy. 
In comparison to handcrafted features, our learned feature with \textit{SVM} gets better results than \textit{3DHOG}, \textit{Cuboids}, and \textit{Gabor3D+HOG3D}, whereas, almost equal performance is achieved when compared to the combination of \textit{HOG}, \textit{HOF}, \textit{MBH}, and \textit{Trajectories} descriptors \cite{Wang2011}, highlighting more discriminative power of our learned features. 
\newline
We adopted the trained model on full \textit{Weizmann} dataset and pre-process it similarly to \textit{KTH}. 
The same \textit{3DPyraNet-F} and \textit{3DPyraNet-F\_M} models were applied. In this case, despite more classes, optimal results were achieved compared to state-of-the-art as shown in Table.\ref{table:KTHWEIZMANCOMPARISON}. \textit{3DPyraNet-F} enhances previous results by $8.09\%$, whereas, \textit{3DPyraNet-F\_M} enhanced it further with an additional $0.14\%$. As compared to combination of $(HOG + HOF + MBH + Trajectories)$, \textit{3DPyraNet-F\_M} have a lower accuracy of $0.87\%$.    
\vspace{-0.5em}
\subsubsection{Dynamic Scene Recognition/Understanding}
In \textit{DSR}, a model has to learn the whole mask rather than a specific portion of the image. \textit{3DPyraNet} weight matrix is of equal size to input image/feature map. Therefore, it could be an ideal case for \textit{DSR} in videos that can be used as a hint in other recognition tasks. The model for these datasets has a bigger input size compared to previous datasets for \textit{AR}, i.e. $80\times100\times13$ resulting in a feature vector of size 33075. 
We considered an overlap of 7 frames, considering a small number of frames compared to previous models \cite{Tran2014a,Derpanis2012,feichtenhofer2014bags}. For instance, \cite{Tran2014a} uses $128\times171\times16$ frames in a clip from which $112\times112\times16$ random crops were extracted for data augmentation. 
\newline
In case of \textit{YUPENN} dataset, our model achieved best accuracy of $96.2134\%$ after 25 epochs. However, it achieves mean accuracy of $93.67\%$ for a one-vs-rest classification. Although, it is better than \cite{feichtenhofer2013spacetime, 6619180, Derpanis2012} by huge margin, still it does not achieve state-of-the-art performance by $5.33\%$ fewer accuracy. One of the reasons could be that the model in \cite{Feichtenhofer2016} combined certain complex pre-processing and feature extraction techniques (PCA, LLC, GMM, IFV, static pooling and their proposed dynamic spacetime pyramid pooling in SPM) that overcomes even previous optimal results provided by deep model \textit{C3D}.  Further, in comparison to \textit{C3D}, possibly, one resides in the fact that \textit{C3D} \cite{Tran2014a} is trained on Sports 1-Million videos dataset \cite{Karpathy2014}, whereas \textit{3DPyraNet-F} is trained on the same small dataset. In addition, \textit{C3D} have high resolution and use data augmentation.
Compare to \textit{C3D} and Imagenet, \textit{3DPyraNet-F} achieves comparable results, i.e. $93.67\%$; a good starting point for future work to test \textit{3DPyraNet-F} on a very large scale dataset. Christoph's et al. model \cite{feichtenhofer2014bags} performance is better than ours by $1.5\%$ (Table.\ref{table:KTHWEIZMANCOMPARISON}), but their result is based on majority voting for video classification whereas, we did individual clip classification. 
\newline
In order to further evaluate the strength of \textit{3DPyraNet-F}, unlike \textit{YUPENN}, we tested it with \textit{MaryLand} dataset that includes camera induced motion. Despite camera motion, we achieved a state-of-the-art accuracy of 94.83\% as shown in Table. \ref{table:KTHWEIZMANCOMPARISON}, representing the discriminative power of the proposed fusion model. \textit{3DPyraNet-F} outperforms state-of-the-art method \cite{Tran2014a} by 7.17\%. Whereas, state-of-the-art model in-terms of \textit{YUPENN} by 14.87\%. 
Classes such as Boiling water, fountain, iceberg collapse, whirlpool shows slight poor results. One of the reasons could be that all the classes contain some similarity, i.e. water, which brings ambiguity that make it hard to classify correctly. 
Table. \ref{table:perClassCompa} and \ref{table:perClassComparison} shows the performance for each class in comparison to current state-of-the-art models. In case of \textit{MaryLand}, we outperformed all the classes other than chaotic traffic and waves. However, in case of \textit{YUPENN} dataset, our model showed poor performance for sky-clouds and windmill farm, while showing comparable results on other categories.  
\vspace{-0.5em}
\subsection{Parameters Reduction}
\vspace{-0.5em}
After the strong success of the Alex ConvNet model with ImageNet dataset \cite{Krizhevsky2012b,Tran2014a}, models became deeper and deeper. Beside accuracy, their trainable parameters also increased. 
The number of parameters is unarguably a substantial issue in application space. It results in hihg memory cost and large size of trained models on the disk \cite{Krizhevsky2012b,Tran2014a,learningConnections15,NINLinCY13}. Separate consideration should be made for the reduction of parameters \cite{7727350}.
\newline
\textit{NiN} \cite{NINLinCY13} highlighted the issue of reducing parameters, but \textit{NiN} achieved it at greater computational cost. \textit{NiN} uses a FC \textit{MLP} as a filter. Recently, this FC \textit{MLP} is made sparse in \textit{CiC} model \cite{Pang2016} while using unshared LC scheme. \textit{CiC} not only reduces parameters but shows better performance than \textit{NiN}. C. Szegedy et al. \cite{SzegedyLJSRAEVR15} uses sparsity reduction complex methodologies over the trained models for refining the \textit{FC} layers becuase most of the parameters are in \textit{FC} layers. 
S. Han et al. model \cite{learningConnections15} learns the connections in each layer instead of weights and then the network is trained again to reduce the number of parameters.
\newline
We compare our model against state-of-the-art \textit{C3D} model in terms of performance and less number of parameters. \textit{C3D} has about $17.5M$ parameters, whereas our models have less than a million parameters (specifically 0.83M parameters in the case of \textit{YUPENN} and \textit{MaryLand} dataset). 
Disk occupancy/usage is almost negligible compared to the model trained by \textit{C3D} as shown in Table. \ref{table:KTHWEIZMANCOMPARISON}; this is of great help in embedded systems and mobile devices where the memory usage is a problem.
\vspace{-0.5em}
\section{Conclusion}
\label{sec:Conc}
\vspace{-0.5em}
A strict pyramidal 3D NN has been proposed that process and learn features from raw input frames of a given video. 
\textit{3DPyraNet}, due to its biologically inspired pyramid structure is a deep model that is capable to learn effective features in fewer layers and less parameters as compared to recent deep competitors, despite camera induced motion. 
It has been shown here that a good architecture can achieve competitive results even with a limited amount of data.
Furthermore, the proposed fusion based model with a linear-SVM classifier for feature learning has achieved competitive results with respect to current best methods on different video analysis benchmarks for \textit{AR} and \textit{DSR}.
In the future, the widespread applicability of \textit{3DPyraNet} and its fusion based variations will be verified by validating it on recent challenging datasets e.g. UCF sports, YouTube action, since the model is aimed to obtain good performances despite the complexity and diversity of the tackled tasks. In addition, as deep models are hard to explain or interpret, the learned weights will be analysed for explainability and interpretability of the model and the decision it take.
%
\bibliographystyle{abbrvnat}
\bibliography{main}
\end{document}